\documentclass[10pt,twocolumn,letterpaper]{article}

\usepackage{cvpr}
\usepackage{times}
\usepackage{epsfig}
\usepackage{graphicx}
\usepackage{amsmath}
\usepackage{amssymb}
\usepackage{multirow}
\usepackage[singlelinecheck=false]{caption}
\usepackage{array, caption, subcaption,tabularx,  ragged2e,  booktabs, threeparttable}

\usepackage[shortlabels]{enumitem}

\usepackage[breaklinks=true,bookmarks=false]{hyperref}
\usepackage{authblk}

\cvprfinalcopy 

\def\cvprPaperID{****} 

\begin{document}

\title{Object Detection in Aerial Imagery}

\author{Dmitry Demidov}
\author{Salem AlMarri}
\author{Rushali Grandhe}
\affil{Mohamed bin Zayed University of Artificial Intelligence}
\affil[ ]{\textit {\{firstname.lastname\}@mbzuai.ac.ae}}

\maketitle

\begin{abstract}
    Object detection in natural images has achieved remarkable results over the years. However, a similar progress has not yet been observed in aerial object detection due to several challenges, such as high resolution images, instances scale variation, class imbalance etc. We show the performance of two-stage, one-stage and attention based object detectors on the iSAID dataset. Furthermore, we describe some modifications and analysis performed for different models - \\
    \textbf{in two stage detector:} introduced weighted attention based FPN, class balanced sampler and density prediction head.\\
    \textbf{in one stage detector:} used weighted focal loss and introduced FPN\\
    \textbf{in attention based detector:} compare single,multi-scale attention and demonstrate effect of different backbones
    \footnote{The pre-trained models and video can be found online: \url{https://mbzuaiac-my.sharepoint.com/:f:/g/personal/dmitry_demidov_mbzuai_ac_ae/EkWESOWeAlZLpme5j_t2Ql8Bcfwn7gkiul_JDgZAJ16OqA?e=ssZjPd}}.\\ Finally, we show a comparative study highlighting the pros and cons of different models in aerial imagery setting.
    

    
\end{abstract}


\section{Introduction}

Object detection is a very popular computer vision task which involves both objects locating and classifying in an image. 
Though this task has been extensively studied for usual natural images, there is limited exploration in the field of aerial object detection. Aerial object detection brings various challenges such as large scale variation, uneven distribution of objects, class imbalance etc. which further makes the detection of objects much more demanding. However, the problem is particularly interesting due to its important applications such as defence, forestry, city planning etc.

In this project we have performed experiments on the iSAID dataset which is a Large-scale Dataset for Object detection and Instance Segmentation in Aerial Images (iSAID) \cite{waqas2019isaid}. It includes 2,806 high-resolution images taken by satellites, and collects 665,451 densely annotated object instances, belonging to a total of 15 classes. 


In this work we, first, perform object detection in aerial images using two-stage, one-stage and attention-base detectors. Owing to the significant variation in the model architectures, it is natural that the same modification may not be suitable for all architectures. Hence, we describe modifications  and analysis performed in each of the 3 models. \textbf{For the two-stage model}, we have used the Faster-RCNN architecture where we introduced modifications such as weighted attention based FPN to better handle scal variation, simple class balanced sampler to handle the inherent long tail distribution and a density prediction head to tackle the large density variation observed in and among different images. \textbf{For the one stage model}, we have used the YOLOv5 model and tried to analyse the performance with the addition of weighted focal loss to counter class imabalnce and FPN in attempt to deal with various scales. \textbf{For attention based detector,} we have used DETR model. We tried to analyse the effect of different attention levels such as single scale, multi scale attention and the impact of different bakbones such as ResNet50,ResNet101 etc. 

And finally, we demonstrate a comparison between each of the considered detectors with respect to aerial object detection.


\section{Methods}

The used iSAID dataset and the architectures we experimented with: two-stage, one stage and attention based object detectors, are discussed in the following subsections. 

\subsection{Dataset}

A Large-scale Dataset for Object Detection and Instance Segmentation in Aerial Images (iSAID) \cite{waqas2019isaid} is the dataset which was specifically developed for the tasks of object detection, semantic and instance segmentation of aerial images. It includes 2,806 high-resolution images taken by satellites, and collects  
665,451 densely annotated object instances, belonging to a total of 15 classes. The iSAID follows the same annotation format used in the popular MS COCO dataset and provides bounding boxes and pixel-level annotations, where each pixel represents a particular class (or absence of any). In order to leverage the limitations of input size of the model used, the original images were cropped into equal overlapping patches with 800x800px resolution.


\begin{figure} [!ht]
    \centering
    \includegraphics[scale = 0.69]{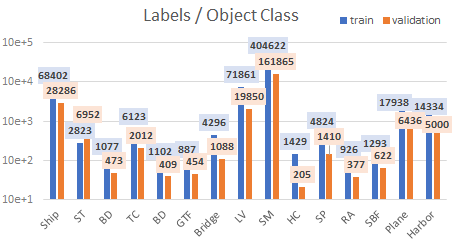}
    \caption{Number of instances per each class in the train and validation sets of the iSAID dataset \cite{waqas2019isaid}. Notations: ST - storage tank, BD - baseball diamond, TC - tennis court, BC - basketball court, GTF - ground track field, LV - large vehicle, SV - small vehicle, HC - helicopter, RA - roundabout, SBF - soccer ball field, SP - swimming pool.}
    \label{fig:labels}
\end{figure}

The dataset has several distinguishing characteristics, such as large number of images with high spatial resolution, large count of labelled instances per image (which might help in learning contextual information), noticeable scene variation depending on an object type, and imbalanced and uneven distribution of objects. One of the most important key characteristics of the dataset is its huge size and scale variation of the objects, providing the instance size range from 10x10px to 300x300px, which is also the case for instances of the same class. A representation for some of the above-mentioned properties can be seen in the figure \ref{fig:labels}.

\subsection{Two stage detector}
Several two stage approaches have been developed for object detection. R-CNN \cite{girshick2014rich} used selective search algorithm to generate bounding box proposals. Fast R-CNN \cite{girshick2015fast} is an extension of R-CNN\cite{girshick2014rich} which enabled end-to-end training using shared convolutional features, thus improving train and inference speed while also increasing detection accuracy. Faster R-CNN \cite{ren2016faster} replaced selective search with Region Proposal Network (RPN) which generates multi-scale and translation-invariant region proposals based on the high-level features of images. It uses RoI pooling to crop the feature maps according to the proposals. This is followed by a small convolution network with classification and regression branch. We have carried out experiments using Faster-RCNN implemented in Detectron2 \cite{wu2019detectron2} framework with ResNet-101 FPN backbone. We introduced some modifications in the Faster-RCNN architecture inorder to address some issues specific to the dataset as explained below.\\
\textbf{Handling scale variation: } Feature pyramid network has been a popular approach used to handle scale variation. It includes a bottom up network and top-down network connected using 1x1 lateral connections. This is followed by a convolutional layer with large kernel to increase receptive field. However, FPN still struggles to deal with huge scale variation. This motivated us to introduce a modification in the FPN so that it can better handle varying size of objects.

It is well known that small objects are more likely to be identified in the higher resolution layer P5, medium size objects in layer P4/P3 and large objects can be identified from the low resolution layers P2/P3 of FPN. This indicates that having access to more information i.e. from high resolution feature maps could be helpful for better dealing with varying sizes of objects.

\begin{figure}[!ht]
    \centering
    \includegraphics[width=0.8\linewidth]{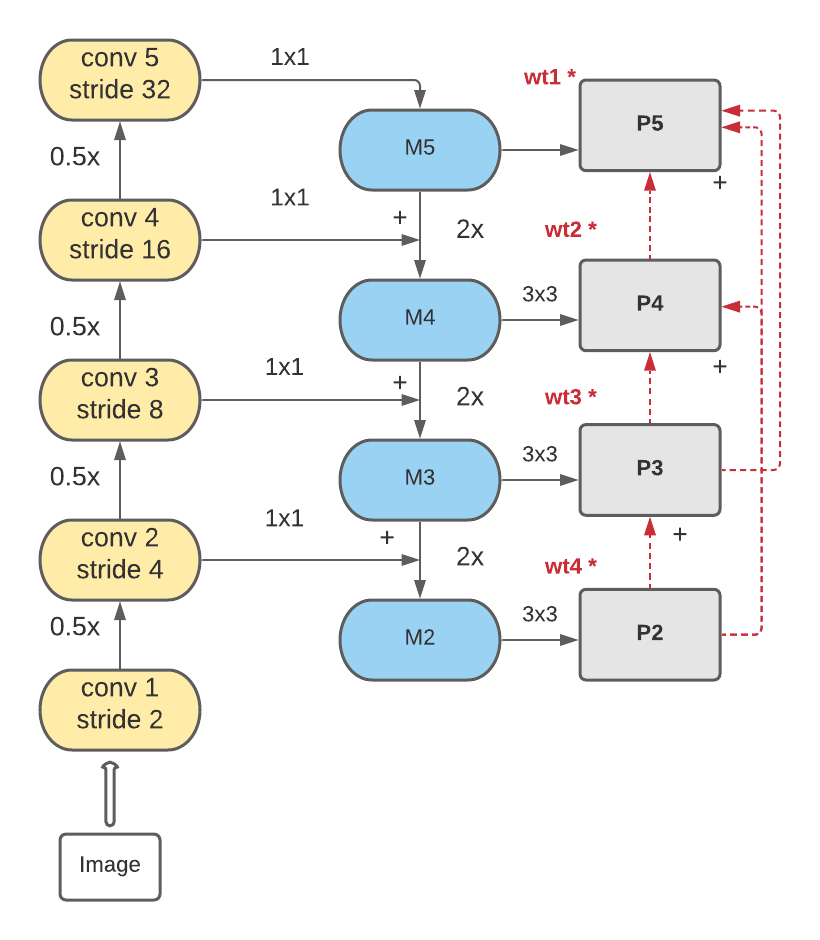}
    \caption{Weighted FPN with channel attention}
    \label{fig:fpn}
\end{figure}

We transformed this intuition such that information from higher resolution feature maps be transferred to lower resolution feature maps. This information has been incorporated in the form of channel attention over the feature maps by drawing inspiration from works such as Squeeze-Excitation blocks \cite{hu2019squeezeandexcitation} and CBAM \cite{woo2018cbam}. However, owing to this change there was an improvement observed in the mAP but it was prominent only for medium and large objects which appears to be logical. Thus, inorder to better deal with the small objects as well, we introduced a small weighting factor for each of the output feature maps as shown in Fig \ref{fig:fpn}, where $wt_4$ $>$ $wt_3$ $>$ $wt_2$ $>$ $wt_1$.

\textbf{Handling class imbalance: } In the iSAID dataset, a lot of class imbalance is observed which also been depicted in Fig \ref{fig:labels}. This corresponds to the problem of long tail distribution where some classes have large number of instances and some classes have very few number of instances. This can in turn cause the model to be biased towards the frequent classes and hence perform worse on the rare classes. We tried to address this issue by implementing a class balanced foreground sampler in the RoI head. Based on the statistics of the dataset, we segregated the classes into 3 different groups i.e. frequent, common and rare and adjusted the number of proposals for each group.\\

\begin{figure}[!ht]
        \centering
        \includegraphics[width=\linewidth]{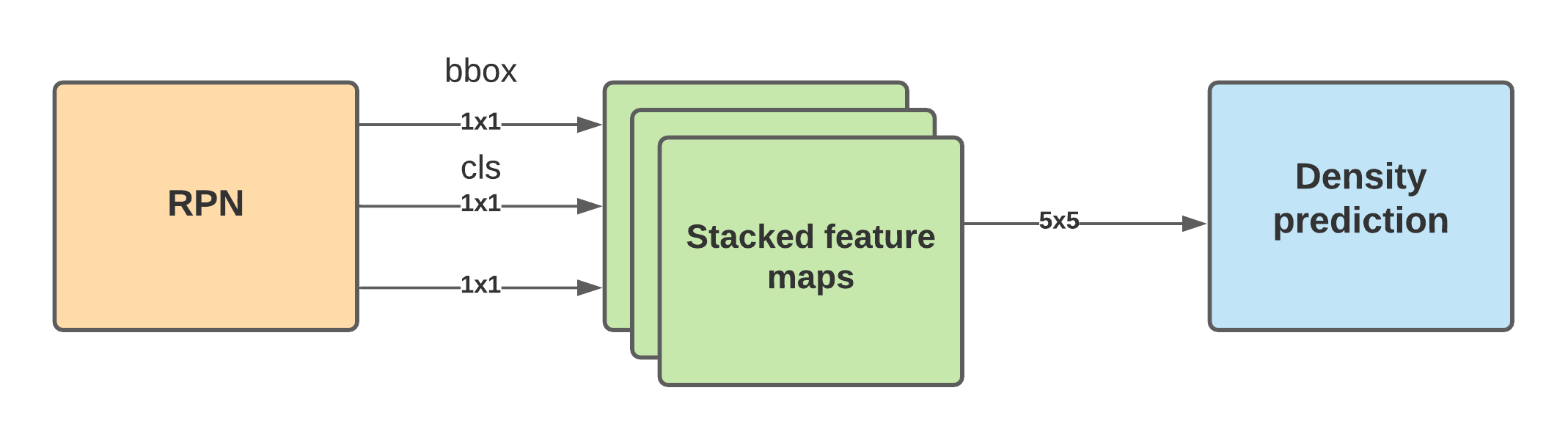}
        \caption{Density prediction head for the Faster R-CNN}
        \label{fig:density}
    \end{figure}

\textbf{Handling uneven distribution/ modelling context: } From the dataset, it can be observed that there is a lot of variation in the distribution of instances within an image as well as across the dataset. This uneven distribution of instances could confuse the model due to the persistent variation. This motivated us to think about modeling the distribution/context about an object instance and introduce the density prediction head in the RPN. The density prediction head was first introduced in the Adapative NMS paper \cite{liu2019adaptive} and was used to capture the surrounding density inorder to adaptively adjust the NMS threshold. However, we use the same density prediction head for a different objective. 
It stacks the classification and bounding box regression branch feature maps along with the output of a 1x1 convolutional layer. This is followed by a convolutional layer with 5x5 kernel which effectively helps capture considerable information in the surroundings of an object as shown in Fig \ref{fig:density}. The corresponding ground truth density is calculated as $d_i$ := max $b_j \epsilon \mathcal{G}$ ,i$\neq$j iou($b_i,b_j$),
where the density of the object i is defined as the max bounding box IoU with other objects in the ground truth set $\mathcal{G}$ and smooth L1 loss is used.
Furthermore, to the best of our knowledge we are the first to employ the density prediction head for aerial imagery.

\subsection{One stage detector}
In one-stage detectors, a one-shot configuration is proposed to replace region proposals, such that classification and regression take place immediately on candidate anchor boxes. With this improvement, architecture is simpler and inference time is more suitable for realistic applications. SSD, YOLO and RetinaNet are examples of one-stage detectors \cite{7780460} \cite{DBLP:journals/corr/LiuAESR15} \cite{DBLP:journals/corr/abs-1708-02002}.YOLOv5 architecture consists of three parts: (1) Backbone: CSPDarknet, (2) Neck: PANet, and (3) Head: Yolo Layer. The images are first input to CSPDarknet for feature extraction, and then fed to PANet for feature fusion. Finally, Yolo layer outputs detection results (class, score, location, size).\\
We introduced some modifications in the YOLOv5 architecture inorder to address some issues specific to the dataset as explained below.\\
\textbf{Focal Loss :} The iSAID dataset consists of multiple object classes which can be grouped based on the number of occurrences per image, certain object classes have higher occurrences rate such as Small vehicles when compared to other classes such as bridge, helicopter, roundabout etc. as shown in figure \ref{fig:labels}. 
For model training, it is expected to have a specific number of training examples per class, with an overall assumption of equal distribution of data, else the model could be biased. In \cite{lin2018focal} a work around for class imbalance is introduced, which is considered as an extension for cross-entropy loss function named as focal loss. There are two adjustable parameters for focal loss, $\gamma$ and $\alpha$. Increasing $\gamma$ value brings model's attention towards object class which are difficult to classify, and increasing $\alpha$ value results in dedicating more weights for object classes with lower annotations. 

\textbf{Feature Pyramid Network :}
Motivated by scale-variation and the ability of FPNs to deal with multiple scales, YOLOv5 architecture was further enhanced with a Feature Pyramid Network. After adding FPN, an mAP of 0.422 was observed on the validation set. 


\subsection{Attention-based detector}

\begin{figure}[!ht]
    \centering
    \includegraphics[width=1.0\linewidth]{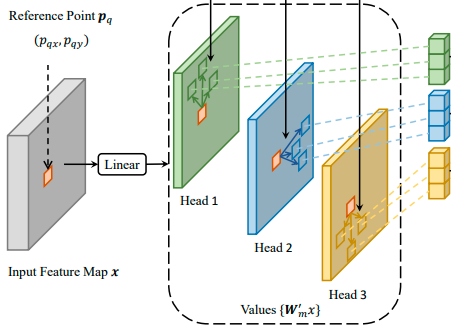}
    \caption{Illustration of the deformable attention module (the aggregation part is not shown) \cite{zhu2021deformable}}
    \label{fig:deformable}
\end{figure}

Another approach considered in this work is called Deformable DETR (Deformable Transformers for End-to-End Object Detection) \cite{zhu2021deformable}, which is an attention-based detector. This recently proposed method attempts to both eliminate the previous need of manually-designed components in object detection while still demonstrating good performance and efficiency.
As can be seen in the figure \ref{fig:deformable}, Deformable DETR combines two popular computer vision techniques, thereby leveraging their advantages in the way that they can be applied for solving drawbacks of each other.

The first component is a transformer-based object detection model, DETR (End-to-End Object Detection with Transformers) \cite{carion2020endtoend}. This quite recently published method solves the object detection task as a direct set prediction problem. The approach provides a streamlined detection pipeline and effectively removes the need for many hand-designed components like a non-maximum suppression procedure or anchor generation that explicitly encode prior knowledge about the task. The main ingredients of this framework, called DEtection TRansformer or simply DETR, are a set-based global loss that forces unique predictions via bipartite matching, and a transformer encoder-decoder architecture. Given a fixed small set of learned object queries, DETR identifies the relations of the objects and the global image context in order to directly output the final set of predictions in parallel. According to the provided by the authors code and various experiments, the model is conceptually simple and efficient, unlike the majority of other modern detectors. DETR demonstrates accuracy and run-time performance on par with the well-established and highly-optimised Faster R-CNN baseline \cite{girshick2015fast}.

The second important part, used to build Deformable DETR, is Deformable Convolutional Networks \cite{dai2017deformable}. It was proved by the authors that the convolutional neural networks (CNNs) with original convolutional layers \cite{cnn} are inherently limited to model geometric transformations due to the fixed geometric structures. To solve this limitation by enhancing the transformation modeling capacity of CNNs, a deformable convolution module was introduced. It is based on the idea of augmenting the spatial sampling locations in the modules with additional offsets and learning the offsets from target tasks, without additional supervision. The new modules, as authors mentioned, can easily replace their plain counterparts in existing CNNs, and can be easily end-to-end trained by standard back-propagation. The high effectiveness of Deformable CNNs was also validated with extensive experiments on such complex vision tasks as object detection and semantic segmentation.

Therefore, the combination of both Deformable convolution, which helps to solve the problem of sparse spatial sampling, and DETR Transformer, which is responsible for relation modeling capability, results in a model that have such advantages as fast convergence, computational feasibility, and memory efficiency.
Moreover, the proposed multi-scale deformable attention architecture attends to only a small set of sampling locations on the feature map pixels, while still having adequate performance, what is presented by the authors as a reasonable replacement for manually-optimised FPN and computationally inefficient full-attention.

Provided in the original paper \cite{zhu2021deformable} results of extensive experiments with Deformable DETR, which was introduced with a goal to mitigate the slow convergence and high complexity issues of DETR, indeed show that on the MS COCO \cite{lin2015microsoft} the benchmark demonstrates its superior effectiveness.
The visualisation of results indicates that the proposed combination of these above-mentioned modules in Deformable DETR looks at extreme points of the object to determine its bounding box, which is similar to the observation in DETR. However, Deformable DETR, more concretely, besides attending to left/right and top/bottom boundaries of the object, also attends to pixels inside the object for predicting its category, which is different to original DETR \cite{carion2020endtoend}.
In addition, compared with its predecessor, Deformable DETR achieves better performance with significantly smaller number of training epochs. This effect is also especially noticeable on small objects, what is, at the same time, can be helpful for the considered iSAID dataset.


\section{Experiments and Results}

\subsection{Faster R-CNN}

\begin{figure}[!ht]
        \centering
        \includegraphics[width=1.1\linewidth]{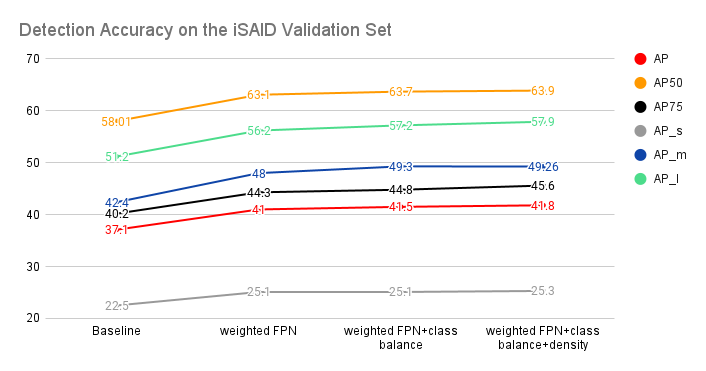}
        \caption{Faster R-CNN modifications ablation study, results on validation set}
        \label{fig:det2 ablation}
    \end{figure}

To further verify the impact of the modifications, we have performed an ablation study and report the results on the iSAID validation set as shown in Fig \ref{fig:det2 ablation}. It can be observed that incrementally adding each modification resulted in better performance indicated by mAP. 
The introduction of weighted FPN with channel attention was able to considerably improve the mAP indicating that using information from higher resolution layers is favorable when dealing with different scale objects. We found that using small weights was beneficial. Emperically, we set $wt_1$=1.5, $wt_2$=2, $wt_3$=2.5, $wt_4$=3.\\
Addition of the class balanced sampler in the RoI head was able to further improve the performance by 0.5 mAP. During training, we tried to ensure that the proposals are selected in a balanced way from each group thus reducing the possibility of bias towards a set of classes. We choose 256 proposals in the RoI head. Usually 25\% (64) proposals are considered as foreground. We ensure that 24 proposals are selected from the rare classes and 20 proposals each from the common and frequent classes.\\
Introducing the density prediction head also led to increase in mAP, however the effect was more pronounced for small objects. This affirms that having knowledge of the context/ density around an instance can be very useful especially for small objects.

All experiments have been carried out with the following settings - batch size of 2, learning rate of 0.0025, momentum of 0.9, weight decay of 0.0001 and $\gamma$ = 0.1 run for 100,000 iterations.


\subsection{YOLOv5}

Inorder to work with the dataset, we had to perform some preprocessing steps -
\subsubsection{Preprocessing: COCO JSON to YOLO TXT}
Datasets dedicated for visual tasks such as object detection or segmentation consist of images and their metadata. The metadata file include label information for each image with annotations of bounding boxes specifying the location of each category found in every image. A widely used format is MS COCO format \cite{10.1007/978-3-319-10602-1_48}. iSAID metadata is provided in COCO JSON format. However, for YOLO it had to be converted into TXT format using an algorithm. Information related to each image was stored in a separate text file titled with respective image name. 
For generating an output json file, a Javascript code was used, which reads each image's name in the directory and replaces it with entry id number.

\subsubsection{Preprocessing: Background Images Reduction}
In YOLO, unlabelled images are used as background images during training to reduce false positives and balance out the weights. This is unlike other models which neglect such images. An experiment was carried out to identify what percentage of background images would yield the highest accuracy and shortest training time. In \ref{tab:table100}, the number of labelled/unlabelled images in training/validation sets are shown as per given iSAID dataset. When YOLOv5L was trained using a resolution of 640, batch size of 16, one epoch at full scale required 31 minutes, but when background images were reduced to 0, training only on labelled images, it took 6.24 minutes per epoch. Moreover, in table \ref{tab:table101}, three models were trained at full-scale, labelled + 10\% unlabelled and labelled images only, where the latter achieved highest accuracy and fastest training time.
 
\begin{table}[h]
\footnotesize
\centering
\begin{tabular}{| c | c | c | c |}
\hline
\textbf{iSAID} & \textbf{Unlabelled} & \textbf{Labelled}\\
\hline
Training Set  & 67703 & 16384  \\
\hline
Validation Set  & 22487 & 6049 \\
\hline
\end{tabular}
\caption{Comparison Table between labelled and unlabelled images}
\label{tab:table100}
\end{table}

\begin{table}[h]
\footnotesize
\centering
\begin{tabular}{| c | c | c | c |}
\hline
\textbf{Images} & \textbf{Training time} & \textbf{Epochs} & \textbf{mAP@.5:.95}\\
\hline
84087 & 179m & 5 & 0.3314 \\
\hline
18018 & 43m & 5 &  0.3448 \\
\hline
16384 & 40m & 5 &  \textbf{0.3811} \\
\hline
\end{tabular}
\caption{Model performance comparison based on background image percentage.}
\label{tab:table101}
\end{table}

We tried playing with $\alpha$ and $\gamma$ parameters in the focal loss function. The bests results can be seen in Table \ref{tab:table102}.

\begin{table}[h]
    \footnotesize
    \centering
    \begin{tabular}{| c | c | c | c |}
    \hline
    \textbf{$\gamma$} & \textbf{$\alpha$} & \textbf{mAP@.5} & \textbf{mAP@.5:.95}\\
    \hline
    1.5 & 0.25 & 0.685 & 0.466 \\
    \hline
    2.0 & 0.25 & 0.689 & 0.469 \\
    \hline
    \end{tabular}
    \caption{Focal loss}
    \label{tab:table102}
\end{table}

YOLOv5 Large architecture (normal), focal loss modified, and FPN modified models were run on iSAID's validation set at resolution of 640, batch size of 16 and epochs of 50 and results are presented in figure \ref{fig:yolo}. In terms of precision and mAP@.5:.95, YOLOv5L scored the highest results when compared to other models. In terms of mAP@.5, the focal loss modified model showed better performance, while FPN modified model scored lowest of all models in all evaluation metrics. The YOLOv5 architecture is highly optimized for object detection tasks and receives consistent updates on uploaded repository. 

\begin{figure}[!ht]
    \centering
    \includegraphics[width=1.1\linewidth]{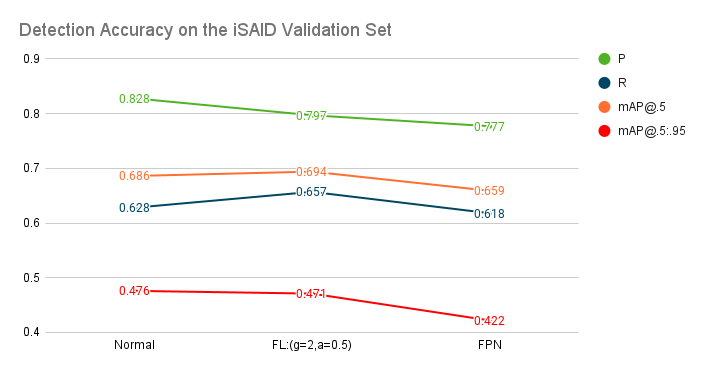}
    \caption{Results for validation set using YOLOv5}
    \label{fig:yolo}
\end{figure}

\subsection{Deformable DETR}

At the moment, due to the recentness of the detection transformer, it has only been tested mainly on the popular, well-studied, and relatively balanced object detection datasets, such as MS COCO \cite{lin2015microsoft}, Pascal \cite{mottaghi_cvpr14}, etc.
In this subsection, we explore the performance of the Deformable DETR and its variations on the iSAID dataset.

If not mentioned another setup, due to the limited time and available computing resources, the following models were trained with the following hyper-parameters: batch size of 6, 15 epochs, learning rate of 2e-4 for encoder-decoder, learning rate of 2e-5 for a backbone, learning rate decays by 0.5 multiplier at each 7th epoch for encoder-decoder and for fully connected classifiers. On average, training time with these parameters is approximately 18-25 hours, depending on the architecture.

\textbf{Singe-Scaled attention:}
First, as a baseline, we chose a single-scaled Deformable DETR model, described in the original paper. It uses an ImageNet pre-trained ResNet-50 convolutional neural network as a backbone for feature extraction. The extracted features from the last convolutional layer with the size of 7x7x2048 are used as an input for the encoder-decoder transformer.
Even this simplified architecture shows decent performance with the mAP value equal to $0.303$, which is quite close to the Faster R-CNN baseline model. Taking into account that the model is trained from scratch for only 15 epochs, we suggest that it is able to reach satisfactory results in case of longer training time. 

\begin{figure}[!ht]
    \centering
    \includegraphics[width=\linewidth]{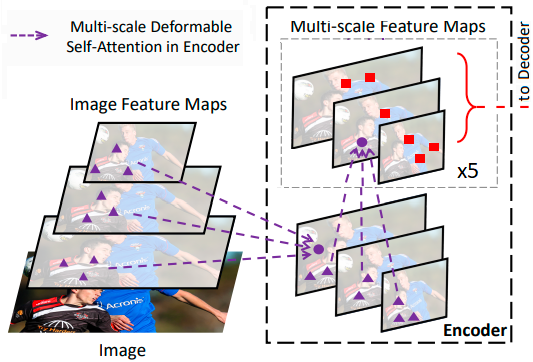}
    \caption{Image feature maps, extracted with a CNN backbone are inputs for a multi-scaled attention module (encoder part); the architecture proposed by \cite{zhu2021deformable}}
    \label{fig:multi-scaled}
\end{figure}

\textbf{Multi-Scaled attention:}
Next, in order to confirm the suggestion, made in \cite{zhu2021deformable} regarding the performance improvement for small and medium-sized objects, when a multi-scaled attention module is used, we also trained this model from scratch.
The multi-scaled encoder transformer module uses Conv3-5 ordinary layers from a CNN backbone, and also uses a Conv5 layer with a stride of 2 applied. All the number of feature channels are projected to 256 to have the same appropriate input size before being fed to the encoder transformer. Architecture of this module can be found in the figure \ref{fig:multi-scaled}.

From the obtained results, one can observe that indeed, the mAP for small and medium objects increased by 20 \%, comparing to the single-scaled architecture. This can be explained by the fact that iSAID dataset has much more corresponding-sized objects, rather than larger ones.
Since this architecture uses considerably more memory to store image features, the batch size value was decreased to 4.

\textbf{ResNet-101 Backbone:}
Another improvement we experimented with is changing the CNN backbone to a larger one, ResNet-101. Although it was not noticeably useful for the usual datasets as COCO and ImageNet where the medium objects are typically still covering significant part of the image frame, we assume that for the iSAID dataset, having larger image resolution, this may help to identify medium instances better. We also expect a certain minor performance increase for the small and large objects, but since the architecture of ResNet-50 and ResNet-101 differ only with the Conv3 layer depth, we assume that the effect should be mostly directed at the medium ones. 

From our experimental observations, this assumption actually takes place, since in case of a single-scale version the accuracy indeed increased by 20 \% for the middle-sized instances, and only by 7-9 \% for the rest of them. In case of the multi-scale attention architecture, the increasing is 10 \% for medium objects, and nearly 3 \% for the rest.

\textbf{Early backbone layers:}
As the next modification we decided to use earlier layers of the CNN backbone. We suggest that implementation of this idea may increase the performance for tiny, small and medium objects. This becomes obvious, that after resizing and passing through the first layers, receptive field of the backbone is for sure smaller, than on the latter layers. Therefore, we suggest that by the moment of reaching the Conv3 layer, the instances with the original resolution smaller than $22x22$ are almost impossible to classify, since they become a nearly 1x1 feature pixel by that layer. This might also explain the fact that the accuracy for small objects is relatively small in case of the considered ordinary Deformable DETR approach.

\begin{figure}[!ht]
    \centering
    \includegraphics[width=\linewidth]{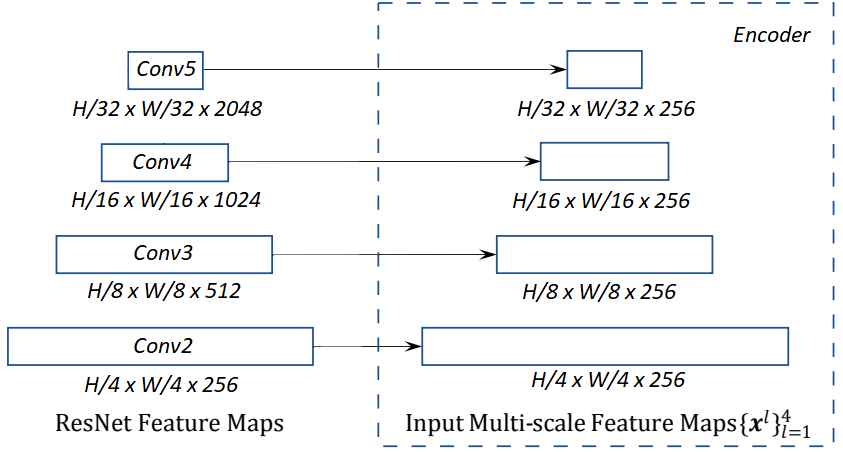}
    \caption{Modified encoder input architecture, connected with the backbone. The Conv2-5 convolutional layers, which outputs are projected to 256 channels, are then used as inputs for the encoder part of multi-scaled attention module.}
    \label{fig:1layer}
\end{figure}

Modified encoder input architecture, shown in the figure \ref{fig:1layer}, now includes the Conv2-5 backbone layers, omitting the last Conv5 layer with a stride of 2 applied previously. Since the size of added Conv2 layer is significantly larger than the size of the removed layer, the resulted model requires for more GPU memory during the training process. Having limited computing resources, we decided to decrease the batch size to 2, which significantly affected training time, increasing it from 0.8 of an hour to 3 hours per epoch.\\


    
    

\textbf{Optimised model:}
Finally, a model with the performance we were able to achieve is a combination of previously mentioned modifications to Deformable DETR. It includes Multi-scale attention module, ResNet-101 as a CNN backbone with earlier layers taken.
More specifically, the model was trained with these hyper-parameters: batch size of 6, 80 epochs, 2e-4 encoder-decoder learning rate, 2e-5 backbone learning rate, learning rate decay by 0.5 at each 7th epoch, and the loss function coefficients optimised for fine-tuning (decreased by 50 \%).

\begin{figure}[!ht]
    \centering
    \includegraphics[width=\linewidth]{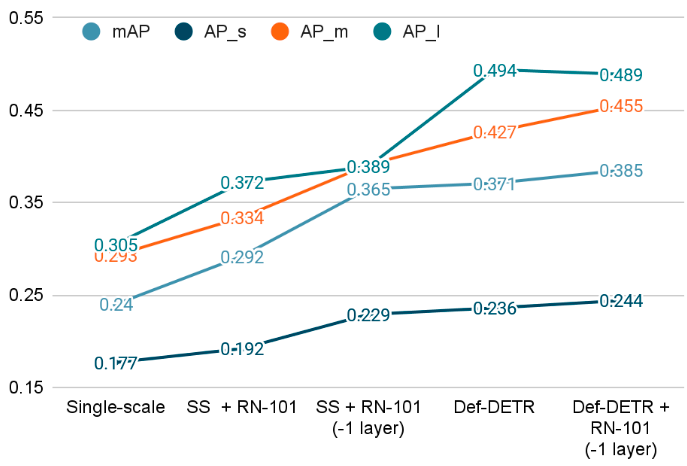}
    \caption{Results on the validation set using different modifications of Deformable DETR. The results for models with denoted as '-1 layer' are expected extrapolated results after training for 30 epochs. Calculation is based on the thorough observations of similar fully-trained for 80 epochs models.}
    \label{fig:yolo}
\end{figure}

\section{Conclusions}

After investigating of the conducted experiments with both original and modified architectures and summarising different performance metrics observed for all three considered approaches, we can provide a model-based conclusion for each of them, mentioning their successes and fails.

Starting with Faster R-CNN we noticed that the architecture shows good precision and recall values on small objects and outstanding results on medium-sized [$P_m: 0.476, R_m:0.684$] (harbor, storage tank, bridge), however it performs quite bad on large ones [$P_l: 0.19$].


Next, YOLOv5, in turn, demonstrates the best precision on small objects [$P_s: 0.46$] (helicopter, large vehicle, baseball diamond) and good values for the small ones, but unacceptably low precision and recall for the large ones [$P_l: 0.13, R_l: 0.25$].

And finally, attention-based Deformable DETR, shows the superior performance indicators for the large objects [$P_l: 0.476, R_l:0.684$] and good values for the medium-sized, however performing worse on the small ones [$P_s: 0.22, R_s: 0.32$].


To conclude, the aerial object detection is certainly a challenging task, providing multiple complex questions to solve. Through our experiments, we tried to analyse the strengths and weaknesses of different types of detectors on the aerial images from iSAID dataset. The observed variety of outcomes and performance of models shows that the object detection problem already can be adequately solved with a certain highly dataset-depended solution, but it still does not have a single approach that can provide a general solution for every case.

{\small
\bibliographystyle{ieee}
\bibliography{egbib}
}

\end{document}